\documentclass{article}
\PassOptionsToPackage{numbers,compress}{natbib}

    \usepackage[preprint]{neurips_2025}

\usepackage{ifthen}
\usepackage{algorithm}
\usepackage{multirow}
\usepackage{booktabs} 
\usepackage[table]{xcolor} 
\usepackage{amsmath}
\usepackage{upgreek} 
\PassOptionsToPackage{numbers,compress}{natbib}
\newboolean{paperfinal}

\usepackage[utf8]{inputenc} 
\usepackage[T1]{fontenc}    
\usepackage{subcaption}
\usepackage{graphicx}
\usepackage{hyperref}       
\usepackage{url}            
\usepackage{booktabs}       
\usepackage{amsfonts}      
\usepackage{nicefrac}       
\usepackage{microtype}
\usepackage{amsmath}
\usepackage{enumitem}
\usepackage{float}
\usepackage{chngcntr}       
\usepackage{titletoc}       
\usepackage{wrapfig}        
\usepackage{tikz}
\usepackage{etoolbox}
\usepackage{tabularx} 
\usepackage{amsthm}
\usepackage{pifont}
\usepackage{algpseudocode}

\newtheorem{assumption}{Assumption}

\title{Shackled Dancing: A Bit-Locked Diffusion Algorithm for Lossless and Controllable Image Steganography}

\author{%
  Tianshuo Zhang \\
  Harbin Engineering University \\
  Harbin, 150001, China \\
  \texttt{zhang.tianhshuo@163.com} \\
  \And
  GaoJia \\
  Harbin Engineering University \\
  Harbin, 150001, China \\
  \texttt{gj944468183@gmail.com} \\
  \AND
  Wenzhe Zhai \\
  Harbin Engineering University \\
  Harbin, 150001, China \\
  \texttt{wenzhezhai@163.com} \\
  \And
  Rui Yann \\
  Harbin Engineering University\\
  Harbin, 150001, China \\
  \texttt{Shu1l0n9@gmail.com} \\
  \And
  Xianglei Xing\thanks{Corresponding author.} \\
  Harbin Engineering University \\
  Harbin, 150001, China \\
  \texttt{xingxl@hrbeu.edu.cn} \\
}

\begin{document}

\maketitle

\begin{abstract}
Data steganography aims to conceal information within visual content, yet existing spatial- and frequency-domain approaches suffer from trade-offs between security, capacity, and perceptual quality. Recent advances in generative models, particularly diffusion models, offer new avenues for adaptive image synthesis, but integrating precise information embedding into the generative process remains challenging. We introduce Shackled Dancing Diffusion, or SD$^2$, a plug-and-play generative steganography method that combines bit-position locking with diffusion sampling injection to enable controllable information embedding within the generative trajectory. SD$^2$ leverages the expressive power of diffusion models to synthesize diverse carrier images while maintaining full message recovery with $100\%$  accuracy. Our method achieves a favorable balance between randomness and constraint, enhancing robustness against steganalysis without compromising image fidelity. Extensive experiments show that SD$^2$ substantially outperforms prior methods in security, embedding capacity, and stability. This algorithm offers new insights into controllable generation and opens promising directions for secure visual communication.
\end{abstract}

\section{Introduction}\label{sec:introfuction}

As digital systems become increasingly pervasive, protecting sensitive data has become a critical challenge. While conventional cryptographic approaches~\cite{hao2023hybrid,gao2023color,zhou2024novel} provide robust security guarantees, their explicit use often signals the presence of valuable information, inadvertently inviting adversarial attention. This has led to growing interest in data steganography~\cite{yin2021reversible}, which enables covert communication by embedding hidden payloads within innocuous carriers, such as images, to evade detection altogether.

Traditional image steganography methods fall into two main categories: spatial-domain and frequency-domain techniques. Spatial-domain approaches—such as Least Significant Bit (LSB) substitution~\cite{chai2020efficient,rahman2023huffman,geetha2021multi,laimeche2020enhancing,kaur2021pvo}—offer high payload capacity but are vulnerable to statistical steganalysis due to fixed embedding patterns~\cite{rahman2023huffman}. On the other hand, frequency-domain methods improve robustness by embedding information in transformed representations~\cite{miri2017adaptive,jeluvsic2022low,lan2023robust}, often leveraging wavelet or Fourier transforms. However, these techniques suffer from reduced payload capacity and potential distortion due to irreversible transformations.

Recent advances in deep learning, particularly in computer vision~\cite{shi2022semantic,fan2025pcpt,zhai2024zero}, have brought new momentum to steganography. Generative models—especially diffusion-based algorithms~\cite{ho2020denoising,liu2024physics3d}—offer powerful tools for high-fidelity, distribution-aligned data generation. Early diffusion-based steganography approaches~\cite{peng2024ldstega,yu2024cross,xu2024image} have demonstrated promising results, yet face notable limitations: either constrained payloads, model dependence, or excessive architectural complexity.

In this work, we propose a novel steganographic algorithm that bridges the strengths of traditional and generative paradigms, , as illustrated in Figure \ref{fig:framework diagram}. By integrating bit-level manipulation from LSB methods, permutation-based encryption, and the generative flexibility of diffusion models, our approach enables high-capacity, secure, and imperceptible data hiding across multiple modalities. The proposed algorithm is named \textbf{Shackled Dancing Diffusion (SD\textsuperscript{2})} to reflect its core intuition: although constraints are imposed during the diffusion process by embedding hidden information, the model still generates diverse and meaningful images. Much like "dancing in shackles," the diffusion process remains expressive and controllable under carefully designed limitations, achieving a balance between generative freedom and steganographic precision. The contributions of this paper can be summarized as follows: 

\begin{figure}[H]
    \centering
    \includegraphics[width=\linewidth]{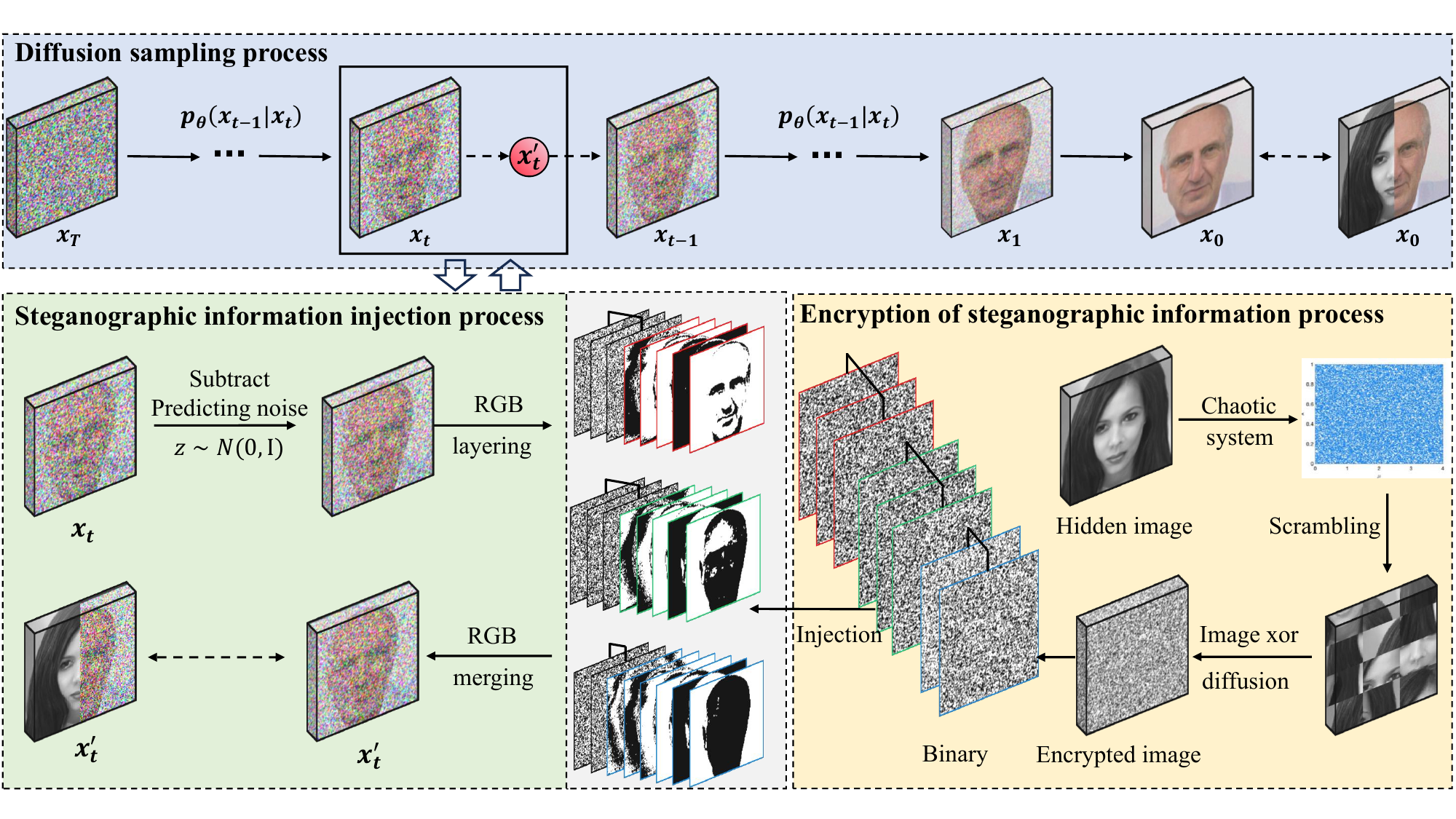}
    \caption{Process framework diagram of SD\textsuperscript{2}.}
    \label{fig:framework diagram}
\end{figure}

\begin{enumerate}[label=\arabic*), leftmargin=16pt, itemsep=0.5em]
    \item We propose SD$^2$, a plug-and-play module that integrates bit-position locking with diffusion sampling injection to enable precise and robust message embedding in generative image synthesis.
    \item We achieve $100\%$ message recovery, ensuring lossless decoding under a wide range of conditions, while preserving high perceptual quality in the generated images.
    \item We empirically demonstrate significant improvements over state-of-the-art methods in terms of embedding capacity, security, and robustness to perturbation.
    \item We provide a new perspective on controllable generation with constrained randomness, bridging generative modeling and secure communication.
\end{enumerate}

The remainder of the paper is organized as follows. Section~\ref{sec:related_works} reviews related work. Section~\ref{sec:Method} details the proposed approach. Section~\ref{sec:experiments} presents experimental results and analysis. Section~\ref{sec:con} concludes with a discussion of limitations and future directions.

\section{Related Works}\label{sec:related_works}

\paragraph{Deep Learning for Image Steganography.}
Recent years have witnessed a growing interest in leveraging deep neural networks for image steganography, significantly advancing both embedding capacity and imperceptibility. Early approaches employed convolutional autoencoders~\cite{zhu2018hidden,zhang2019steganogan} to jointly optimize the encoder-decoder pipeline, enabling end-to-end hiding and extraction of secret data. More sophisticated methods, such as HiDDeN~\cite{zhu2018hidden}, adopted adversarial training to resist statistical and steganalytic attacks, while architectures like SteganoGAN~\cite{zhang2019steganogan} further improved visual quality and payload robustness through GAN-based frameworks. With the rise of attention mechanisms, transformer-based steganography~\cite{wang2022deep} has emerged to exploit global context in both cover and secret domains. Recently, generative models have opened up new directions for covert communication. Meanwhile, diffusion models have introduced an alternative paradigm where secret messages can be seamlessly integrated into the denoising process~\cite{wei2023generative,yu2024cross}, offering high-capacity, high-invisibility solutions. For example, CRoSS~\cite{yu2024cross} leverages pre-trained models such as Stable Diffusion to enable a controllable, robust, and secure steganographic framework without requiring additional training. However, most existing steganographic methods tend to optimize for a single objective—such as capacity, imperceptibility, or robustness—making it challenging to achieve a well-balanced performance across all criteria.

\paragraph{Denoising Diffusion Probabilistic Model.}

\begin{wrapfigure}{r}{0.5\textwidth}
    \centering
    \includegraphics[width=\linewidth]{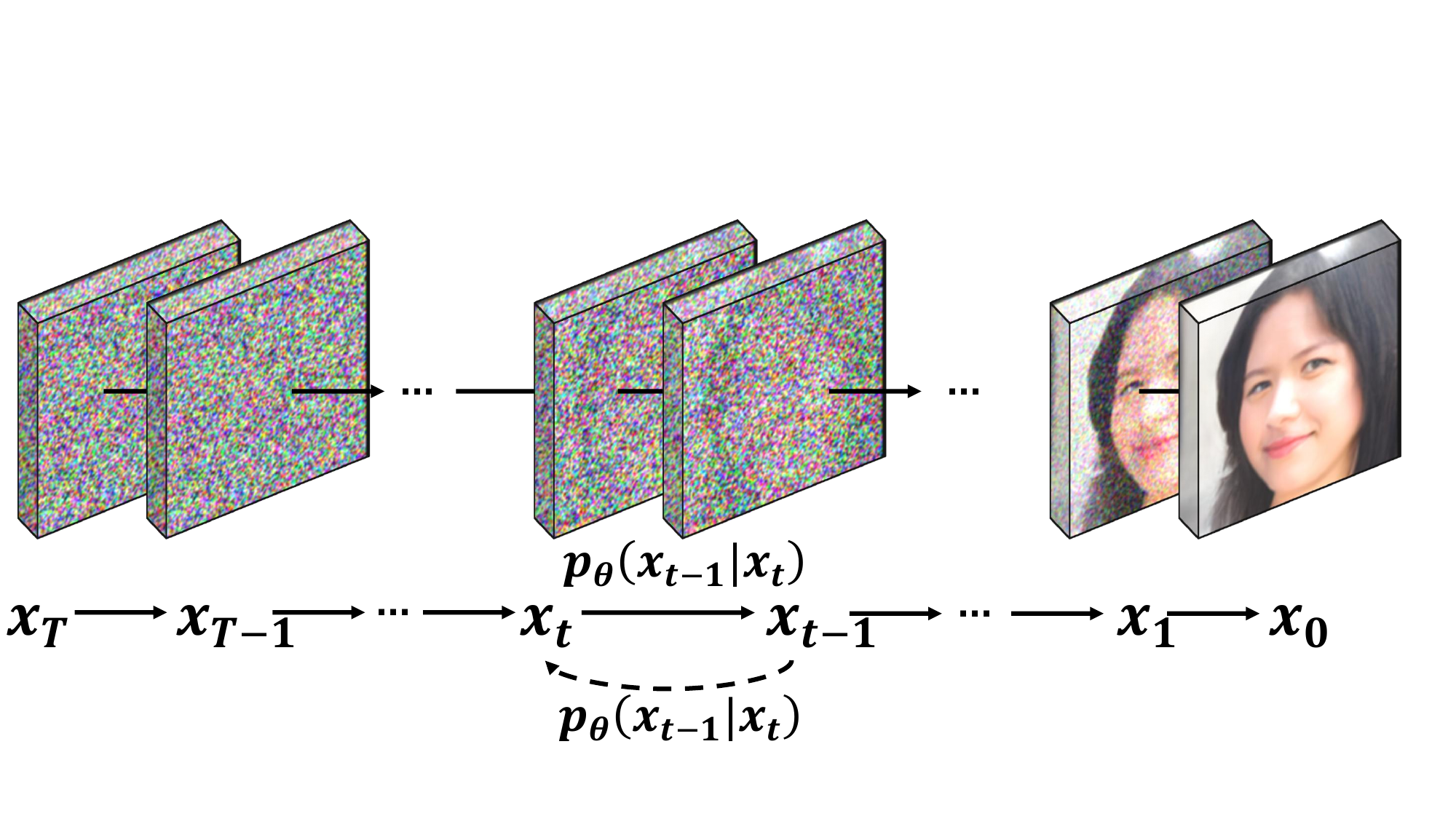}
    \caption{Schematic diagram of diffusion model.}
    \label{fig:Markov chain}
\end{wrapfigure}

Diffusion models have emerged as a dominant paradigm for high-quality image synthesis, outperforming GAN-based methods in both fidelity and diversity~\cite{ho2020denoising,rombach2022high,dhariwal2021diffusion}. As shown in Figure \ref{fig:Markov chain}, these models generate images from pure noise by learning to reverse a Markov noise process. DDPMs~\cite{ho2020denoising} first demonstrated the potential of diffusion in image generation, inspiring subsequent variants such as improved DDPMs and classifier-free guidance~\cite{ho2022classifier,yu2022scaling}, which introduced more effective sampling and conditional generation. Latent Diffusion Models (LDMs)~\cite{rombach2022high} significantly reduce computational cost by operating in a compressed latent space, while maintaining high generation quality, becoming the foundation for models like Stable Diffusion. Recent works have leveraged pre-trained diffusion backbones for controllable generation via text prompts~\cite{rombach2022high}, sketches~\cite{kwon2022diffusion}, or depth maps~\cite{kim2022dag}, enabling broad applications across visual domains. Additionally, Score Distillation Sampling (SDS)~\cite{poole2022dreamfusion} introduces a loss formulation that allows optimizing input structures, such as 3D representations or scene layouts, by aligning diffusion scores with generated outputs, catalyzing its use in inverse graphics and 3D generation pipelines~\cite{seo2023let,zou2024triplane}. The emergence of diffusion models has opened up new directions for image steganography, offering more flexible and high-fidelity frameworks for concealing information within generative processes.

\section{Method}
\label{sec:Method}

Diffusion models inherently exhibit a one-to-many mapping from a Gaussian prior to a diverse set of semantically coherent outputs. Introducing stochastic processes into diffusion models leads to a key insight: among the vast number of potential outputs that conform to the empirical distribution of complex images, there exists a subset that can satisfy specific constraints. These constraints are characterized by fixed bit values at designated pixel locations. As illustrated in Figure~\ref{fig:Potential Steganographic Targets}, our analysis reveals a compelling conclusion: within the range of meaningful images generated by diffusion models, certain outputs may subtly embed additional information. This hidden data is seamlessly integrated into the image's structure, becoming an intrinsic part of it, such that no noticeable anomalies can be detected upon superficial examination of the generated image.

Leveraging this property, we propose \textbf{SD$^2$}, a diffusion-based steganographic algorithm illustrated in Figure~\ref{fig:framework diagram}. The method encodes a secret message into a pseudo-random binary stream using a chaotic mapping, and embeds bits during specific denoising steps by fixing selected bit positions within the 8-bit binary representation of chosen pixels.

As formalized in Algorithm~\ref{alg:stego_sampling}, embedding occurs at timesteps $t \in T_I$ by manipulating the least significant bits of $\mathbf{x}_t$, the latent at time $t$, using the denoising model $\epsilon_\theta$ and diffusion parameters $\alpha_t$, $\overline{\alpha}_t$, and $\sigma_t$. 


The mathematical proofs provided in the appendix~\ref{sec:Mathematical Proof}, along with the empirical results in Section~\ref{sec:experiments}, collectively validate the feasibility of our proposed approach. To better illustrate the complete pipeline of SD$^2$, we present a detailed walkthrough using the task of hiding a grayscale image of size $m \times n$ as a representative example. The main procedure can be summarized as follows:

\begin{algorithm}[H]
\caption{Steganographic Information Embedding during Diffusion Sampling}
\label{alg:stego_sampling}
\begin{algorithmic}[1]
\State $\mathbf{x}_T \sim \mathcal{N}(0, \mathbf{I})$ \Comment{Initialize latent with Gaussian noise}
\For{$t = T$ to $1$}
    \If{$t > 1$}
        \State $\mathbf{z} \sim \mathcal{N}(0, \mathbf{I})$
    \Else
        \State $\mathbf{z} = \mathbf{0}$
    \EndIf
    \State $\mathbf{x}_{t-1} \gets \frac{1}{\sqrt{\alpha_t}} \left( \mathbf{x}_t - \frac{1 - \overline{\alpha_t}}{\sqrt{1 - \overline{\alpha_t}}} \epsilon_\theta(\mathbf{x}_t, t) \right) + \sigma_t \mathbf{z}$ \Comment{Reverse denoising}
    
    \If{$t \in T_I$}
        \State $b \gets \text{Next bit from message stream}$
        \State Select channel $c \in \{R, G, B\}$ and pixel $(i,j)$
        \State $v \gets \mathbf{x}_t^{(c)}[i,j]$ \Comment{Locked pixel bits}
        \State $v^{\text{bin}} \gets \text{To8BitBinary}(v)$
        \State Choose index $k \in \{4,5,6,7\}$
        \State $v^{\text{bin}}[k] \gets b$
        \State $\mathbf{x}_t^{(c)}[i,j] \gets \text{From8BitBinary}(v^{\text{bin}})$
    \EndIf
\EndFor
\State \Return $\mathbf{x}_0$
\end{algorithmic}
\end{algorithm}

\paragraph{Steganography Image Encryption Process.}
Assume that the image $I$, which is concealed, is a grayscale image of dimensions $m \times n$, with a total of $m \times n$ pixels. To achieve effective information hiding, a one-dimensional chaotic sequence $X_{\text{chaos}}$, of the same length $m \times n$, is generated based on the chaotic system~\cite{pak2017new} defined by Eq. (\ref{eq:chaos}). Each element of this chaotic sequence corresponds to a pixel in the image $I$, providing high randomness and unpredictability for subsequent operations.

\begin{equation}
x_{n+1} = F(\mu, x_n, k) = F_{chaos}(\mu, x_n) \times G(k) - \text{floor}\left(F_{chaos}(\mu, x_n) \times G(k)\right).
\label{eq:chaos}
\end{equation}
where $x_n$ is the sequence generated by the chaotic map, $n$ represents the number of iterations, $F_{chaos}(\mu, x_n)$ denotes the traditional one-dimensional chaotic map, while d refers to the extended chaotic map. Under the conditions of $k \in [8, 20]$ and $\mu \in [0, 10]$, the extended chaotic system demonstrates superior chaotic performance compared to the original map.

Next, $X_{\text{chaos}}$ is sorted in ascending order, and the positions of each original element in the sorted sequence are recorded, resulting in an index sequence $O$ (Ordered sequence). The image $I$ is then converted into a one-dimensional vector $I_{\text{vec}}$ (with a length of $m \times n$), and this vector is rearranged according to the index sequence $O$, producing a scrambled image $I_2$. This scrambling step disrupts the spatial arrangement of pixels, significantly enhancing the system's resistance to cropping attacks. Even if parts of the image are damaged, the integrity of the concealed information is largely preserved, thereby improving the robustness of the system.

To adapt the chaotic sequence $X_{\text{chaos}}$ to the pixel value range of grayscale images, it is mapped to the interval $[0, 255]$, resulting in a mapped random sequence $M$, referred to as the "mapped sequence." Subsequently, each pixel value of the scrambled image $I_2$ is subjected to an XOR operation with the corresponding element of the $M$ sequence, generating the encrypted image $I_3$. This XOR process not only ensures that the original information is embedded into seemingly random values, thereby strengthening the security of information hiding, but also guarantees the reversibility of the process, facilitating decryption and information recovery.

By combining the randomness of the chaotic sequence with the efficiency of pixel scrambling and XOR operations, the proposed method significantly enhances the security of information hiding while improving the system's robustness against complex attacks.

\paragraph{Steganographic Information Embedding Process in Carrier Image Generation.}
To generate an $m \times n$ RGB color image that matches the dimensions of the target hidden image, a pretrained diffusion model and its parameters are utilized. The diffusion model follows a stepwise denoising process modeled as a Markov chain, progressively transitioning from a highly noisy state $x_t$ to the original noise-free image state $x_0$. At each step, the trained parameters predict the noise present in the current state $x_t$ and update the latent variables accordingly.

\begin{wrapfigure}{r}{0.5\textwidth}
    \centering
    \includegraphics[width=\linewidth]{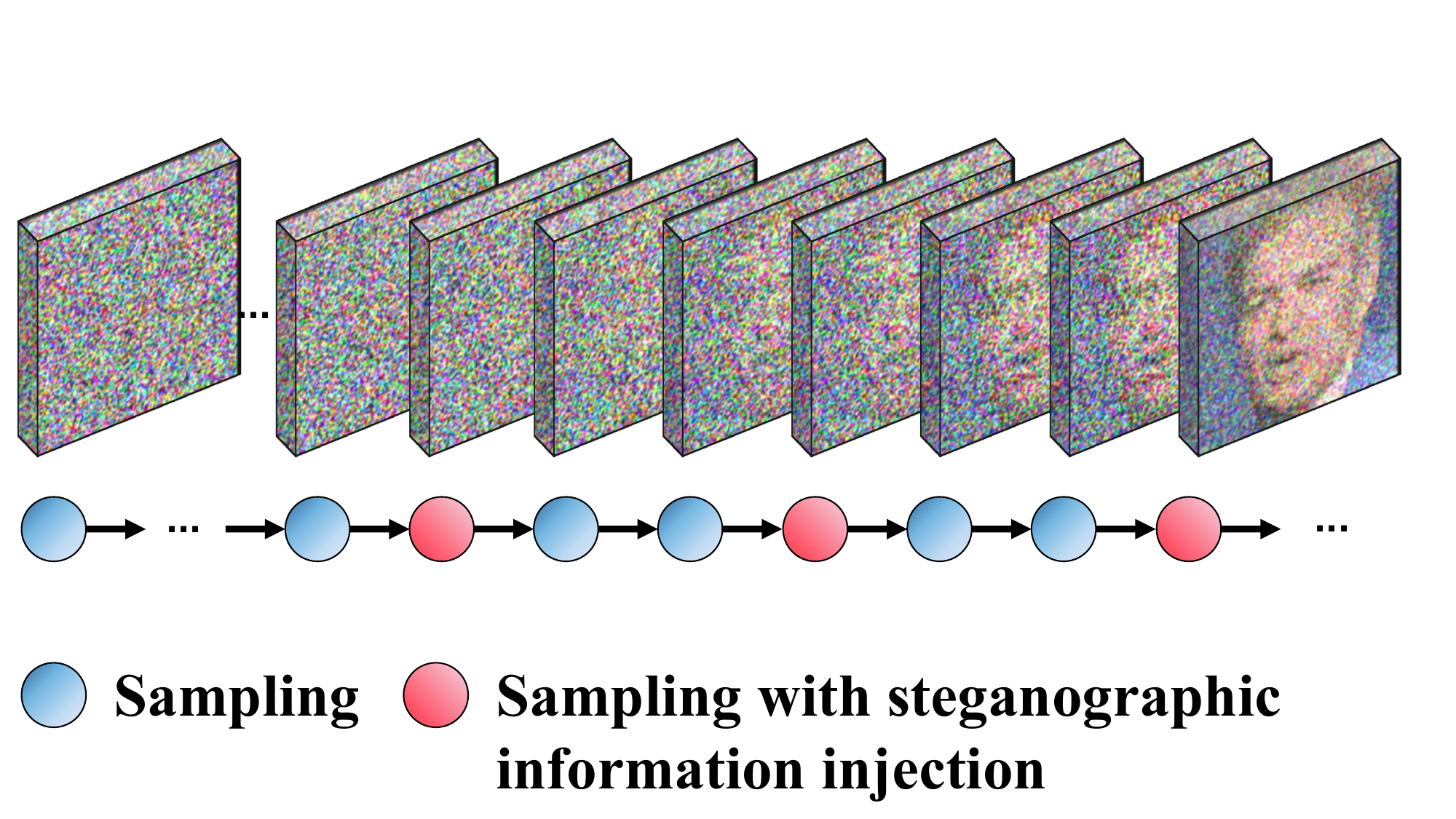}
    \caption{Schematic diagram of the steganography information injection process.}
    \label{fig:steganography information injection points}
\end{wrapfigure}

As shown in Figure \ref{fig:steganography information injection points}, at specific predefined timesteps $t$, a custom operation referred to as bit locking is introduced to securely embed the encrypted grayscale image $I_3$ into the evolving RGB image. Specifically, for each pixel in the $m \times n$ encrypted grayscale image, its value is converted into an 8-bit binary representation. These 8 bits are then partitioned into three groups: the first three bits, the middle three bits, and the final two bits. These bit groups are fixed into the least significant bits of the RGB channels at the state $x_t$, as follows: the first three bits are embedded into the LSBs of the red (R) channel, the middle three bits are embedded into the LSBs of the green (G) channel, and the final two bits are embedded into the LSBs of the blue (B) channel.

Given the human visual system's heightened sensitivity to changes in the blue channel, only minimal modifications are made to the blue channel to preserve the imperceptibility of the embedding process while ensuring the integrity of the hidden information. After the bit locking operation, the modified state $x_t$ is reintroduced into the subsequent denoising steps of the diffusion process for further evolution. The timesteps where bit locking is applied are collectively referred to as $T_i$ (Intervention timesteps). This approach ensures the secure embedding of secret information while maintaining the overall quality of the RGB image.

By incorporating this method, we effectively encode secret information into RGB images while preserving their visual consistency and quality, thereby achieving an efficient and imperceptible steganographic process.


\subsection{Experimental Setup}

\paragraph{Implementation Details.}
\label{Implementation Details}
The hyperparameters are set as follows: $\mu = 3.9$, $r = 0.6$, $k = 14$. The extended chaotic system used in the experiment is Chebyshev chaos~\cite{pak2017new}. All experiments were conducted using the PyTorch framework on NVIDIA RTX 4090 GPU. In the experimental setup, we select every 100 time steps after $T_I$ reaches 1000 for steganographic injection, with additional injections during the final 5 time steps. Specifically, starting from step 1000, we perform bit locking every 100 steps, i.e., at $T_I = 1000, 1100, 1200, \ldots$. Furthermore, in the final stage, we specifically conduct steganographic injections during the last 5 consecutive time steps. This sampling strategy not only effectively reduces computational burden but also avoids excessive interference with the diffusion process, which could otherwise prevent the generation of meaningful images. The evaluation metrics used in our experiments are detailed in Appendix ~\ref{sec:Metric Definitions}.

\paragraph{Datasets.}
We adopt the $128 \times 128$ facial images from the FFHQ-CelebA-HQ dataset~\cite{karras2018progressive}, as discussed in Literature ~\cite{saharia2022image}, as the baseline for generating diffusion models. During the sampling process, we apply the method proposed in ~\cite{saharia2022image}.

\subsection{Experimental results}

\section{Experiments and Results}
\label{sec:experiments}

\begin{figure}[ht]
\centering
\begin{subfigure}[b]{0.095\textheight}
\includegraphics[width=\textwidth]{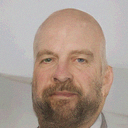}
\caption{}
\end{subfigure}
\hfill
\begin{subfigure}[b]{0.095\textheight}
\includegraphics[width=\textwidth]{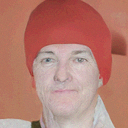}
\caption{}
\end{subfigure}
\hfill
\begin{subfigure}[b]{0.095\textheight}
\includegraphics[width=\textwidth]{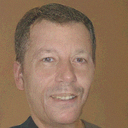}
\caption{}
\end{subfigure}
\hfill
\begin{subfigure}[b]{0.095\textheight}
\includegraphics[width=\textwidth]{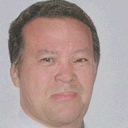}
\caption{}
\end{subfigure}
\hfill
\begin{subfigure}[b]{0.095\textheight}
\includegraphics[width=\textwidth]{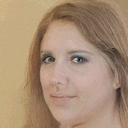}
\caption{}
\end{subfigure}
\hfill
\begin{subfigure}[b]{0.095\textheight}
\includegraphics[width=\textwidth]{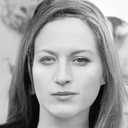}
\caption{}
\end{subfigure}

\vspace{4pt}
\begin{subfigure}[b]{0.095\textheight}
\includegraphics[width=\textwidth]{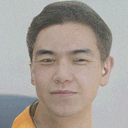}
\caption{}
\end{subfigure}
\hfill
\begin{subfigure}[b]{0.095\textheight}
\includegraphics[width=\textwidth]{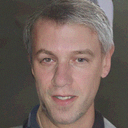}
\caption{}
\end{subfigure}
\hfill
\begin{subfigure}[b]{0.095\textheight}
\includegraphics[width=\textwidth]{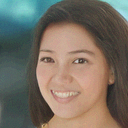}
\caption{}
\end{subfigure}
\hfill
\begin{subfigure}[b]{0.095\textheight}
\includegraphics[width=\textwidth]{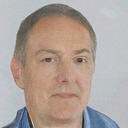}
\caption{}
\end{subfigure}
\hfill
\begin{subfigure}[b]{0.095\textheight}
\includegraphics[width=\textwidth]{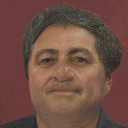}
\caption{}
\end{subfigure}
\hfill
\begin{subfigure}[b]{0.095\textheight}
\includegraphics[width=\textwidth]{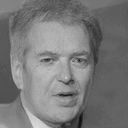}
\caption{}
\end{subfigure}

\vspace{4pt}

\begin{subfigure}[b]{0.095\textheight}
\includegraphics[width=\textwidth]{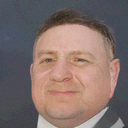}
\caption{}
\end{subfigure}
\hfill
\begin{subfigure}[b]{0.095\textheight}
\includegraphics[width=\textwidth]{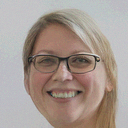}
\caption{}
\end{subfigure}
\hfill
\begin{subfigure}[b]{0.095\textheight}
\includegraphics[width=\textwidth]{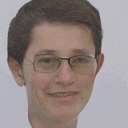}
\caption{}
\end{subfigure}
\hfill
\begin{subfigure}[b]{0.095\textheight}
\includegraphics[width=\textwidth]{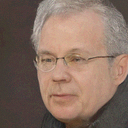}
\caption{}
\end{subfigure}
\hfill
\begin{subfigure}[b]{0.095\textheight}
\includegraphics[width=\textwidth]{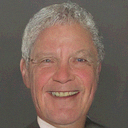}
\caption{}
\end{subfigure}
\hfill
\begin{subfigure}[b]{0.095\textheight}
\includegraphics[width=\textwidth]{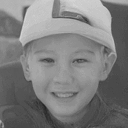}
\caption{}
\end{subfigure}

\begin{subfigure}[b]{0.095\textheight}
\includegraphics[width=\textwidth]{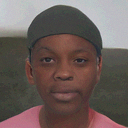}
\caption{}
\end{subfigure}
\hfill
\begin{subfigure}[b]{0.095\textheight}
\includegraphics[width=\textwidth]{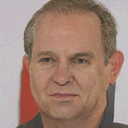}
\caption{}
\end{subfigure}
\hfill
\begin{subfigure}[b]{0.095\textheight}
\includegraphics[width=\textwidth]{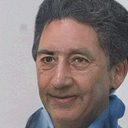}
\caption{}
\end{subfigure}
\hfill
\begin{subfigure}[b]{0.095\textheight}
\includegraphics[width=\textwidth]{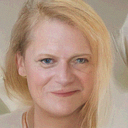}
\caption{}
\end{subfigure}
\hfill
\begin{subfigure}[b]{0.095\textheight}
\includegraphics[width=\textwidth]{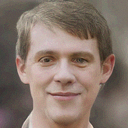}
\caption{}
\end{subfigure}
\hfill
\begin{subfigure}[b]{0.095\textheight}
\includegraphics[width=\textwidth]{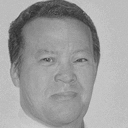}
\caption{}
\end{subfigure}

\caption{Partial steganographic embedding results. (a--e): carriers, (f): hidden image; (g--k): carriers, (l): hidden image; (m--q): carriers, (r): hidden image; (s--w): carriers, (x): hidden image.}

\label{fig:Steganographic Embedding Results}
\end{figure}

\paragraph{Steganography Results.}
As shown in Figure~\ref{fig:Steganographic Embedding Results}, the carrier images remain visually natural and artifact-free after steganographic injection. Notably, the generated face images show no perceptible correlation with the hidden content, demonstrating strong imperceptibility and security. These results confirm that our method effectively embeds information while preserving visual quality and resisting detection.

\begin{table}[ht]
    \centering
    \caption{Comparison of payload capacity (in bits per pixel, bpp) across different steganographic methods.}
    \label{table:payload-capacity}
    \renewcommand{\arraystretch}{1.2}
    \begin{tabularx}{0.95\linewidth}{X c c}
        \toprule
        \textbf{Method} & \textbf{Payload Level} & \textbf{Payload (bpp)$\uparrow$} \\
        \midrule
        Li \textit{et al}.\cite{li2024robust}, Chen\textit{et al}.\cite{chen2022cost}, Li\textit{et al}.\cite{li2023constructing}, Tang\textit{et al}.\cite{tang2024joint}, Zhang\textit{et al}.\cite{zhang2023steganography} & \multirow{4}{*}{\textcolor{red}{Low}} & 0.50 \\
        Pramanik\textit{et al}.\cite{pramanik2023adaptive} & & 0.75 \\
        Lan\textit{et al}.\cite{lan2023robust} & & 1.00 \\
        Chai\textit{et al}.\cite{chai2020efficient}, Rahman\textit{et al}.\cite{rahman2023huffman}, Geetha\textit{et al}.\cite{geetha2021multi}, Laimeche\textit{et al}.\cite{laimeche2020enhancing}, kaur\textit{et al}.\cite{kaur2021pvo} & & 2.00 \\
        \midrule
        Pramanik\textit{et al}.\cite{patwari2023image} & \multirow{2}{*}{\textcolor{orange}{Middle}} & 3.12 \\
        Yin\textit{et al}.\cite{yin2021reversible} & & 3.50 \\
        \midrule
        \cite{wei2022generative}, \cite{su2024stegastylegan} & \multirow{3}{*}{\textcolor{green!50!black}{High}} & 4.00 \\
        Tan\textit{et al}.\cite{tan2021channel} & & \textbf{5.00} \\
        \rowcolor{gray!20}
        \textbf{SD$^2$ (Ours)} & & 4.00 \\
        \bottomrule
    \end{tabularx}
\end{table}

\paragraph{Embedding Capacity and Extraction Accuracy.}
To assess the upper limits of embedding capacity, we progressively increased the payload from 1 to 4 bpp by applying varying bit-locking strategies during sampling. 

\begin{figure}[htbp]
\centering
\begin{subfigure}[b]{0.095\textheight}
\includegraphics[width=\textwidth]{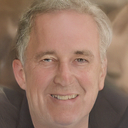}
\caption{}
\end{subfigure}
\hfill
\begin{subfigure}[b]{0.095\textheight}
\includegraphics[width=\textwidth]{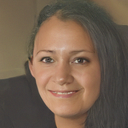}
\caption{}
\end{subfigure}
\hfill
\begin{subfigure}[b]{0.095\textheight}
\includegraphics[width=\textwidth]{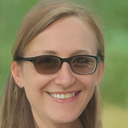}
\caption{}
\end{subfigure}
\hfill
\begin{subfigure}[b]{0.095\textheight}
\includegraphics[width=\textwidth]{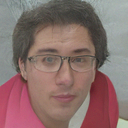}
\caption{}
\end{subfigure}
\hfill
\begin{subfigure}[b]{0.095\textheight}
\includegraphics[width=\textwidth]{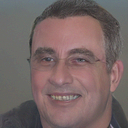}
\caption{}
\end{subfigure}
\hfill
\begin{subfigure}[b]{0.095\textheight}
\includegraphics[width=\textwidth]{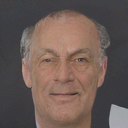}
\caption{}
\end{subfigure}

\vspace{4pt}
\begin{subfigure}[b]{0.095\textheight}
\includegraphics[width=\textwidth]{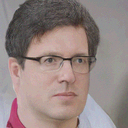}
\caption{}
\end{subfigure}
\hfill
\begin{subfigure}[b]{0.095\textheight}
\includegraphics[width=\textwidth]{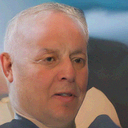}
\caption{}
\end{subfigure}
\hfill
\begin{subfigure}[b]{0.095\textheight}
\includegraphics[width=\textwidth]{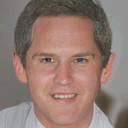}
\caption{}
\end{subfigure}
\hfill
\begin{subfigure}[b]{0.095\textheight}
\includegraphics[width=\textwidth]{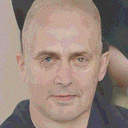}
\caption{}
\end{subfigure}
\hfill
\begin{subfigure}[b]{0.095\textheight}
\includegraphics[width=\textwidth]{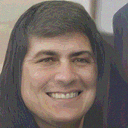}
\caption{}
\end{subfigure}
\hfill
\begin{subfigure}[b]{0.095\textheight}
\includegraphics[width=\textwidth]{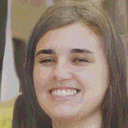}
\caption{}
\end{subfigure}

\caption{Steganographic images under varying embedding capacities. (a)--(c): 1 bpp; (d)--(f): 2 bpp; (g)--(i): 3 bpp; (j)--(l): 4 bpp.}
\label{fig:different embedding capacities}
\end{figure}

\begin{table}[ht]
    \centering
    \begin{minipage}{0.51\textwidth}  
        \centering
        \caption{Accuracy (\%) comparison of different methods on $128 \times 128$ images at varying capacities.}
        \label{table:Accuracy}
        \renewcommand{\arraystretch}{1.2}
        \resizebox{\textwidth}{!}{  
        \begin{tabular}{c r r r >{\columncolor{gray!20}}r}
            \toprule
            \textbf{ACC (\%) } & \textbf{\cite{tan2021channel}} & \textbf{\cite{wei2022generative}} & \textbf{\cite{su2024stegastylegan}} & \textbf{SD$^2$ (Ours)} \\
            \midrule
            @1 bpp & 98.14 & 97.53 & 97.75 & \textbf{100.00} \\
            @2 bpp & 94.92 & 81.61 & 97.41 & \textbf{100.00} \\
            @4 bpp & 89.12 & 70.14 & 94.25 & \textbf{100.00} \\
            \bottomrule
        \end{tabular}
        }
    \end{minipage} \hfill
    \begin{minipage}{0.48\textwidth}
        \centering
        \caption{Comparison of different methods in terms of PSNR, SSIM, BER, and robustness.}
        \label{table_4}
        \begin{tabular}{lccc}
            \toprule
            \textbf{Method} & \textbf{PSNR$\uparrow$} & \textbf{SSIM$\uparrow$} & \textbf{BER$\downarrow$} \\
            \midrule
            \cite{zhu2018hidden} & 36.61 & 0.922 & 24.72 \\
            \cite{hayes2017generating} & 35.54 & 0.914 & 26.29 \\
            \cite{zhang2019steganogan} & 40.47 & 0.971 & 1.43 \\
            \cite{zhang2019steganogan} & 42.38 & 0.988 & 1.95 \\
            \cite{jing2021hinet} & 47.43 & 0.993 & 0.47 \\
            \cite{lan2023robust} & 48.41 & 0.996 & \textbf{0.00} \\
            \rowcolor{gray!20}
            \textbf{SD$^2$ (Ours)} & - & \textbf{1.000} & \textbf{0.00} \\
            \bottomrule
        \end{tabular}
    \end{minipage}
\end{table}

As shown in Figure~\ref{fig:different embedding capacities}, the diffusion model consistently produces high-quality, semantically coherent images, even at 4 bpp, demonstrating strong error correction capabilities under extreme constraints. Table~\ref{table:payload-capacity} presents a comparison of embedding capacity between our approach and representative steganographic methods from recent years. 

For clarity, we categorize methods into three groups based on payload capacity: low ($\leq 2$ bpp), medium ($2\text{--}3$ bpp), and high ($> 3$ bpp). SD$^2$ achieves a competitive payload capacity, placing it firmly within the high-capacity category. While it does not attain the absolute maximum capacity reported, it offers a substantial advantage over other high-capacity methods~\cite{tan2021channel,wei2022generative,su2024stegastylegan}. As shown in Table \ref{table_4}, the carrier images are generated jointly with the hidden content, and thus inherently represent the embedded results; they can therefore be considered equivalent to the post-embedding images.

As shown in Table~\ref{table:Accuracy}, existing approaches with high capacity~\cite{tan2021channel,wei2022generative,su2024stegastylegan} fail to achieve reliable information recovery, often resulting in non-zero bit error rates. In contrast, SD$^2$ enables perfect message extraction, demonstrating a superior balance between capacity and reliability. Taken together, these results highlight the effectiveness of SD$^2$ in achieving both high embedding performance and robust recoverability.

\begin{figure}[H]
\centering
\begin{subfigure}[b]{0.1\textwidth}
\centering
\begin{minipage}[c][1.8\textwidth][c]{\textwidth}
\centering
\textbf{Carrier image}
\end{minipage}
\label{fig:sub1}
\end{subfigure}
\hspace{0.05cm} %
\begin{subfigure}[b]{0.15\textwidth}
\includegraphics[width=\textwidth]{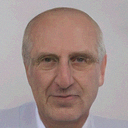}
\caption{}
\end{subfigure}
\hspace{0.05cm} %
\begin{subfigure}[b]{0.15\textwidth}
\includegraphics[width=\textwidth]{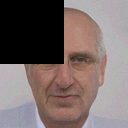}
\caption{}
\end{subfigure}
\hspace{0.05cm} %
\begin{subfigure}[b]{0.15\textwidth}
\includegraphics[width=\textwidth]{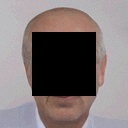}
\caption{}
\end{subfigure}
\hspace{0.05cm} %
\begin{subfigure}[b]{0.15\textwidth}
\includegraphics[width=\textwidth]{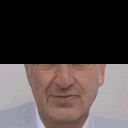}
\caption{}
\end{subfigure}
\hspace{0.05cm} %
\begin{subfigure}[b]{0.15\textwidth}
\includegraphics[width=\textwidth]{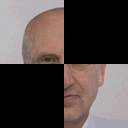}
\caption{}
\end{subfigure}

\vspace{0.05cm} 

\begin{subfigure}[b]{0.1\textwidth}
\centering
\begin{minipage}[c][1.8\textwidth][c]{\textwidth}
\centering
\textbf{Extracted image}
\end{minipage}
\label{fig:sub2}
\end{subfigure}
\hspace{0.05cm} %
\begin{subfigure}[b]{0.15\textwidth}
\includegraphics[width=\textwidth]{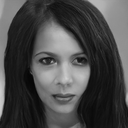}
\caption{}
\end{subfigure}
\hspace{0.05cm} %
\begin{subfigure}[b]{0.15\textwidth}
\includegraphics[width=\textwidth]{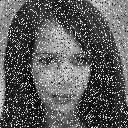}
\caption{}
\end{subfigure}
\hspace{0.05cm} %
\begin{subfigure}[b]{0.15\textwidth}
\includegraphics[width=\textwidth]{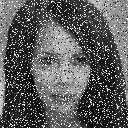}
\caption{}
\end{subfigure}
\hspace{0.05cm} %
\begin{subfigure}[b]{0.15\textwidth}
\includegraphics[width=\textwidth]{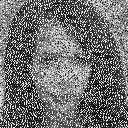}
\caption{}
\end{subfigure}
\hspace{0.05cm} %
\begin{subfigure}[b]{0.15\textwidth}
\includegraphics[width=\textwidth]{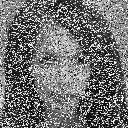}
\caption{}

\end{subfigure}

\caption{Robustness Test of Steganographic Extraction Under Cropping: Each pair of rows shows: (top) the carrier image, and (bottom) the corresponding extracted hidden image.}
\label{fig:Robustness}
\end{figure}

\paragraph{Robustness.}
To assess robustness, we apply cropping operations of varying sizes and positions to the carrier images and evaluate the recoverability of hidden content. As shown in Figure~\ref{fig:Robustness}, our method retains the main structure of the stego image and recovers substantial information even when up to $50\%$ of the image is removed. These results demonstrate strong resilience to cropping attacks and confirm the method's ability to preserve functionality under partial image loss.

\begin{figure}[H]
\centering
\begin{subfigure}[b]{0.15\textwidth}
\includegraphics[width=\textwidth]{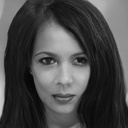}
\caption{}
\end{subfigure}
\hspace{0.05cm} %
\begin{subfigure}[b]{0.15\textwidth}
\includegraphics[width=\textwidth]{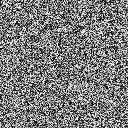}
\caption{}
\end{subfigure}
\hspace{0.05cm} %
\begin{subfigure}[b]{0.15\textwidth}
\includegraphics[width=\textwidth]{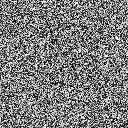}
\caption{}
\end{subfigure}
\hspace{0.05cm} %
\begin{subfigure}[b]{0.15\textwidth}
\includegraphics[width=\textwidth]{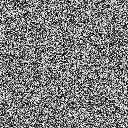}
\caption{}
\end{subfigure}
\hspace{0.05cm} %
\begin{subfigure}[b]{0.15\textwidth}
\includegraphics[width=\textwidth]{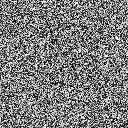}
\caption{}
\end{subfigure}

\caption{Key Sensitivity in Steganographic Extraction Results: (a) shows the result with the correct key; (b) with $\mu + 10^{-15}$; (c) with $r + 10^{-14}$; (d) with $\mu - 10^{-15}$; (e) with $r - 10^{-14}$.}
\label{fig:Key Sensitivity}
\end{figure}

\paragraph{Key Sensitivity and Key Space Analysis.}
To evaluate the security of our method, we conduct a key sensitivity test, as shown in Figure~\ref{fig:Key Sensitivity}. When using the correct key, the hidden image is successfully recovered. However, even minute perturbations in parameters $\mu$ and $r$ (on the order of $10^{-15}$ and $10^{-14}$) result in complete extraction failure, revealing no meaningful information. This confirms the method's strong key dependency and resistance to unauthorized access. Moreover, the key space, determined by the parameters $\mu \in [0, 10]$, $r \in [0, 1]$, and $k \in [8, 20]$ with a precision of $10^{-14}$, exceeds $2^{139}$, providing substantial defense against brute-force attacks~\cite{chai2020efficient}.

\begin{figure}[H]
\centering
\begin{subfigure}[b]{0.12\textwidth}
\centering
\begin{minipage}[c][1.5\textwidth][c]{\textwidth}
\centering
\textbf{Li \textit{et al.}\\ ~\cite{li2024robust}}
\end{minipage}
\label{fig:sub3}
\end{subfigure}
\hfill
\begin{subfigure}[b]{0.13\textwidth}
\includegraphics[width=\textwidth]{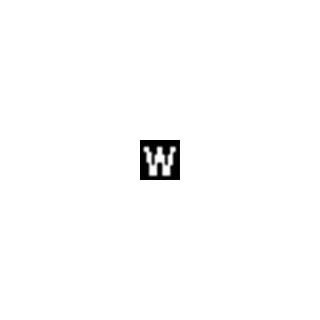}
\caption{}
\end{subfigure}
\hspace{0.05cm} %
\begin{subfigure}[b]{0.13\textwidth}
\includegraphics[width=\textwidth]{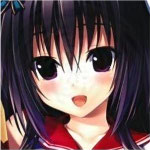}
\caption{}
\end{subfigure}
\hspace{0.05cm} %
\begin{subfigure}[b]{0.13\textwidth}
\includegraphics[width=\textwidth]{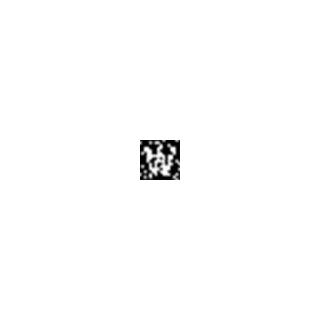}
\caption{}
\end{subfigure}
\begin{subfigure}[b]{0.13\textwidth}
\includegraphics[width=\textwidth]{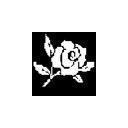}
\caption{}
\end{subfigure}
\hspace{0.05cm} %
\begin{subfigure}[b]{0.13\textwidth}
\includegraphics[width=\textwidth]{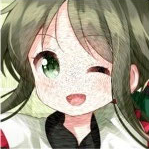}
\caption{}
\end{subfigure}
\hspace{0.05cm} %
\begin{subfigure}[b]{0.13\textwidth}
\includegraphics[width=\textwidth]{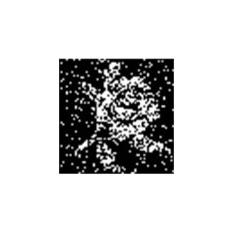}
\caption{}
\end{subfigure}

\vspace{0.05cm} 

\begin{subfigure}[b]{0.12\textwidth}
\centering
\begin{minipage}[c][1.5\textwidth][c]{\textwidth}
\centering
\textbf{SD$^2$\\This work}
\end{minipage}
\label{fig:sub4}
\end{subfigure}
\hfill
\begin{subfigure}[b]{0.13\textwidth}
\includegraphics[width=\textwidth]{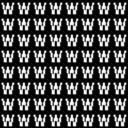}
\caption{}
\end{subfigure}
\hspace{0.05cm} %
\begin{subfigure}[b]{0.13\textwidth}
\includegraphics[width=\textwidth]{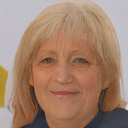}
\caption{}
\end{subfigure}
\hspace{0.05cm} %
\begin{subfigure}[b]{0.13\textwidth}
\includegraphics[width=\textwidth]{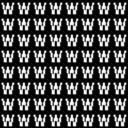}
\caption{}
\end{subfigure}
\begin{subfigure}[b]{0.13\textwidth}
\includegraphics[width=\textwidth]{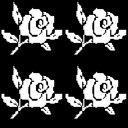}
\caption{}
\end{subfigure}
\hspace{0.05cm} %
\begin{subfigure}[b]{0.13\textwidth}
\includegraphics[width=\textwidth]{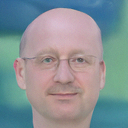}
\caption{}
\end{subfigure}
\hspace{0.05cm} %
\begin{subfigure}[b]{0.13\textwidth}
\includegraphics[width=\textwidth]{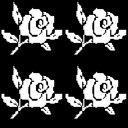}
\caption{}
\end{subfigure}

\begin{subfigure}[b]{0.12\textwidth}
\centering
\begin{minipage}[c][1.5\textwidth][c]{\textwidth}
\centering
\textbf{CRoSS\\ ~\cite{yu2024cross}}
\end{minipage}
\label{fig:sub5}
\end{subfigure}
\hfill
\begin{subfigure}[b]{0.13\textwidth}
\includegraphics[width=\textwidth]{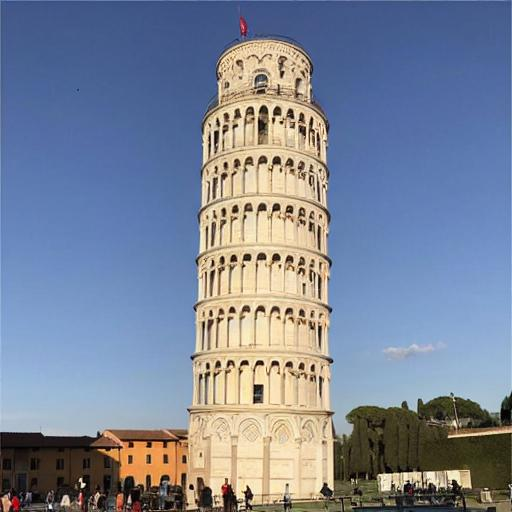}
\caption{}
\end{subfigure}
\hspace{0.05cm} %
\begin{subfigure}[b]{0.13\textwidth}
\includegraphics[width=\textwidth]{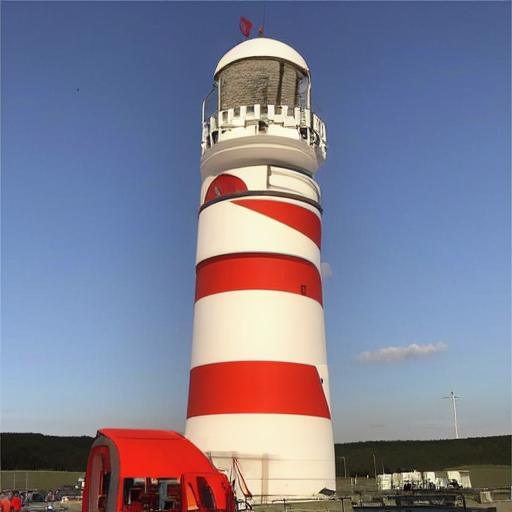}
\caption{}
\end{subfigure}
\hspace{0.05cm} %
\begin{subfigure}[b]{0.13\textwidth}
\includegraphics[width=\textwidth]{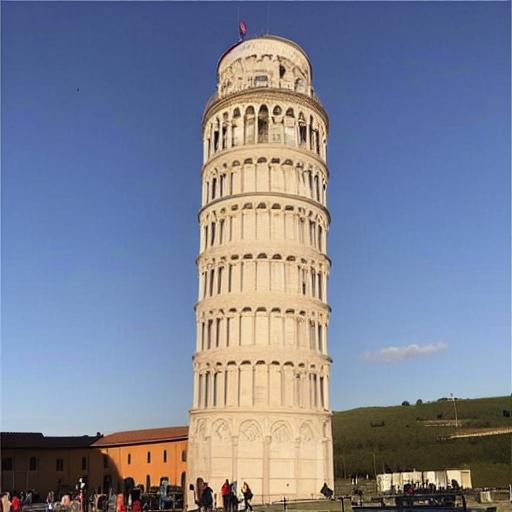}
\caption{}
\end{subfigure}
\begin{subfigure}[b]{0.13\textwidth}
\includegraphics[width=\textwidth]{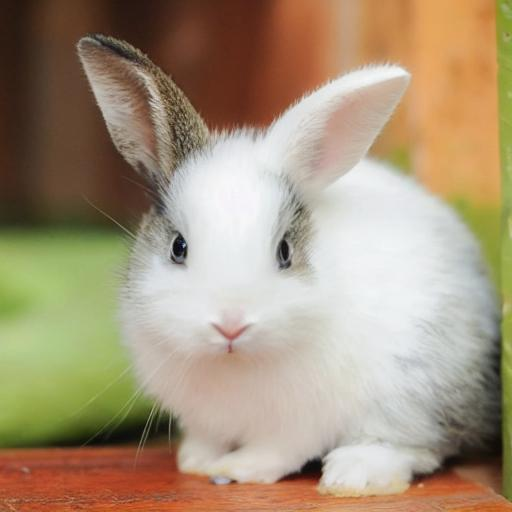}
\caption{}
\end{subfigure}
\hspace{0.05cm} %
\begin{subfigure}[b]{0.13\textwidth}
\includegraphics[width=\textwidth]{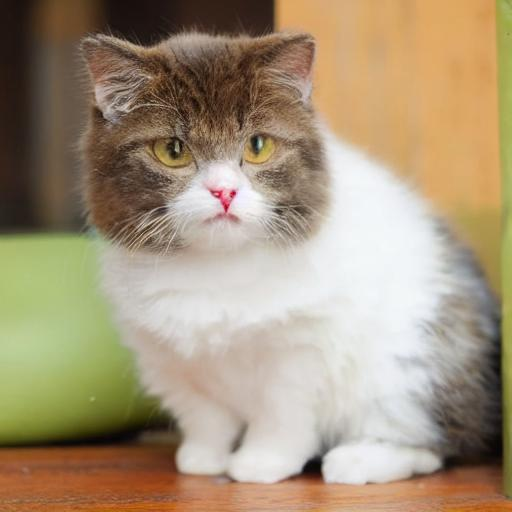}
\caption{}
\end{subfigure}
\hspace{0.05cm} %
\begin{subfigure}[b]{0.13\textwidth}
\includegraphics[width=\textwidth]{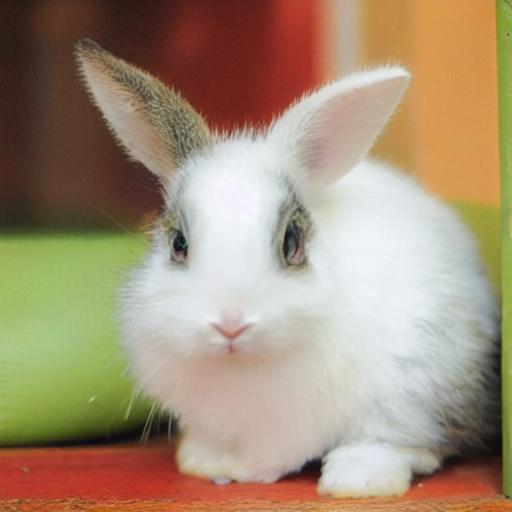}
\caption{}
\end{subfigure}

\vspace{0.05cm} 

\begin{subfigure}[b]{0.12\textwidth}
\centering
\begin{minipage}[c][1.5\textwidth][c]{\textwidth}
\centering
\textbf{SD$^2$\\This work}
\end{minipage}
\label{fig:sub6}
\end{subfigure}
\hfill
\begin{subfigure}[b]{0.13\textwidth}
\includegraphics[width=\textwidth]{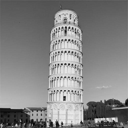}
\caption{}
\end{subfigure}
\hspace{0.05cm} %
\begin{subfigure}[b]{0.13\textwidth}
\includegraphics[width=\textwidth]{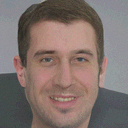}
\caption{}
\end{subfigure}
\hspace{0.05cm} %
\begin{subfigure}[b]{0.13\textwidth}
\includegraphics[width=\textwidth]{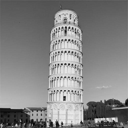}
\caption{}
\end{subfigure}
\begin{subfigure}[b]{0.13\textwidth}
\includegraphics[width=\textwidth]{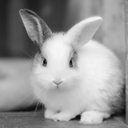}
\caption{}
\end{subfigure}
\hspace{0.05cm} %
\begin{subfigure}[b]{0.13\textwidth}
\includegraphics[width=\textwidth]{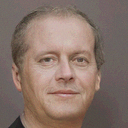}
\caption{}
\end{subfigure}
\hspace{0.05cm} %
\begin{subfigure}[b]{0.13\textwidth}
\includegraphics[width=\textwidth]{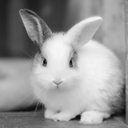}
\caption{}
\end{subfigure}

\caption{Comparison with \cite{yu2024cross} and \cite{li2024robust}. Columns 1-3: Secret image, embedded cover image, extracted info; 
Columns 4-6: Another set of secret image, embedded cover image, and extracted info.}
\label{fig:Comparison of steganographic performance}
\end{figure}

\paragraph{Comparative Results with SOTA Methods.}
As illustrated in Figure~\ref{fig:Comparison of steganographic performance}, we compare our proposed method against two representative steganographic approaches: CRoSS~\cite{yu2024cross}, which is also diffusion-based, and the color transformation method by Li \textit{et al.}~\cite{li2024robust}. For a fair comparison, all methods encode the same secret image content.

\begin{table}[ht]
    \caption{Comparison of different methods in terms of whether the carrier image is generated and embedded, whether the embedded data is fully reversible, and robustness.}
    \label{table:Comparison of different methods}
    \centering
    \begin{tabular}{lcccccc}
        \toprule
        Method & Source  & Generative & Capacity & Fully Reversible & Robustness \\
        \midrule
        \cite{tan2021channel} & TNSE 2021 & {\color{red!70!black}\ding{55}} & \textcolor{green!50!black}{High} & {\color{red!70!black}\ding{55}} & Steganalyzers \\
        \cite{lan2023robust} & AAAI 2023 & {\color{red!70!black}\ding{55}} & \textcolor{red!90!black}{Low} & {\color{green!50!black}\ding{51}} & JPEG compression \\
        \cite{li2023constructing} & TMM 2023 & {\color{red!70!black}\ding{55}} & \textcolor{red!90!black}{Low} & {\color{green!50!black}\ding{51}} & - \\
        \cite{zhang2023steganography} & TDSC 2023 & {\color{green!50!black}\ding{51}} & \textcolor{red!90!black}{Low} & {\color{red!70!black}\ding{55}} & Steganalyzers \\
        \cite{li2024robust} & TCSVT 2024 & {\color{red!70!black}\ding{55}} & \textcolor{red!90!black}{Low} & {\color{red!70!black}\ding{55}} & Steganalyzers / noise \\
        \cite{tang2024joint} & TIFS 2024 & {\color{red!70!black}\ding{55}} & \textcolor{red!90!black}{Low} & {\color{red!70!black}\ding{55}} & Steganalyzers \\
        \cite{yu2024cross} & NeurIPS 2024 & {\color{green!50!black}\ding{51}} & \textcolor{red!90!black}{Low} & {\color{red!70!black}\ding{55}} & Noise \\
        \cite{su2024stegastylegan} & AAAI 2024 & {\color{red!70!black}\ding{55}} & \textcolor{green!50!black}{High} & {\color{red!70!black}\ding{55}} & JPEG compression / noise \\
        \cite{zhou2025improved} & TCSVT 2025 & {\color{green!50!black}\ding{51}} & \textcolor{red!90!black}{Low} & {\color{red!70!black}\ding{55}} & - \\ 
        \rowcolor{gray!20}
        SD$^2$ & Our paper & {\color{green!50!black}\ding{51}} & \textcolor{green!50!black}{High} & {\color{green!50!black}\ding{51}} & Cropping \\
        \bottomrule
    \end{tabular}
\end{table}

CRoSS produces container images that are visually similar to the original secret image, which significantly increases the risk of information leakage. Furthermore, its decoding process is key-dependent and limited to reconstructing semantic content associated with the key, thereby restricting its expressive capacity. In contrast, our method generates visually independent container images, enhancing concealment and supporting more flexible representations of secret information.

Compared to Li \textit{et al.}, our method allows for enlarging the hidden image content by up to $64\times$ and $4\times$, respectively. While their method relies on a $256 \times 256$ container, ours operates effectively with only a $128 \times 128$ container and achieves higher reconstruction fidelity.

These results collectively demonstrate the advantages of our approach in terms of security, embedding capacity, and adaptability. A comprehensive evaluation across four key dimensions—container quality, embedding capacity, recoverability, and robustness—is summarized in Table~\ref{table:Comparison of different methods}. Full journal names of related works are listed in Table~\ref{tab:source_fullnames} in Appendix~\ref{sec:Metric Definitions}.

\section{Conclusion}
\label{sec:con}

We introduce SD$^2$, a diffusion-based steganographic algorithm that embeds information directly into the generative process via a stochastic bit-position locking mechanism. By harnessing the inherent denoising and redundancy-capturing properties of diffusion models, SD$^2$ achieves high-capacity, carrier-free, and precisely decodable message embedding. It outperforms conventional spatial- and frequency-domain techniques in both payload and flexibility, while also surpassing existing deep learning approaches with guaranteed $100\%$ lossless extraction. Unlike prior methods that rely on natural carrier images, SD$^2$ synthesizes visually plausible stego images from scratch, supporting high-fidelity generation even at 4 bpp and demonstrating robustness under severe perturbations, including up to $50\%$ image cropping. Its generality further extends to multimodal inputs such as text and audio, mapped into binary form for unified embedding. Despite these strengths, SD$^2$ currently requires careful bit-position selection and timestep scheduling, which may limit scalability in unconstrained environments. Additionally, while the method ensures high perceptual quality, fine-grained control over semantics remains limited. Future work will explore more structured conditioning, tighter information-theoretic guarantees, and deployment in real-world steganographic scenarios.

\begin{ack}
This work was supported in part by the National Natural Science Foundation of China under Grant 62076078 and in part by the Chinese Association for Artificial Intelligence (CAAI)-Huawei MindSpore Open Fund under Grant CAAIXSJLJJ-2020-033A.
\end{ack}

\bibliographystyle{plainnat}
\bibliography{references}

\newpage\appendix

\section{Metric Definitions}
\label{sec:Metric Definitions}

This section provides definitions and interpretations of the quantitative metrics used in our evaluation, including both steganographic and image quality criteria.

\textbf{Bits Per Pixel (bpp):}
\begin{equation}
\text{bpp} = \frac{|\mathcal{M}|}{H \times W}
\end{equation}
Here, $|\mathcal{M}|$ is the length (in bits) of the embedded message, and $H \times W$ is the resolution of the generated image. This metric reflects the information density per pixel.

\textbf{Accuracy (acc):}
\begin{equation}
\text{acc} = \frac{1}{|\mathcal{M}|} \sum_{i=1}^{|\mathcal{M}|} \mathbf{1} \{ \hat{m}_i = m_i \}
\end{equation}
This measures the bitwise recovery rate of the embedded message. A higher value indicates more accurate retrieval from the generated image.

\textbf{Peak Signal-to-Noise Ratio (PSNR):}
\begin{equation}
\text{PSNR} = 10 \cdot \log_{10} \left( \frac{L^2}{\text{MSE}} \right)
\quad \text{with} \quad
\text{MSE} = \frac{1}{HW} \sum_{i=1}^{H} \sum_{j=1}^{W} (x_{ij} - \hat{x}_{ij})^2
\end{equation}
$L$ denotes the maximum pixel intensity (typically 255). PSNR quantifies image distortion; higher values imply better fidelity.

\textbf{Structural Similarity Index Measure (SSIM):}
\begin{equation}
\text{SSIM}(x, \hat{x}) =
\frac{(2\mu_x \mu_{\hat{x}} + C_1)(2\sigma_{x\hat{x}} + C_2)}
     {(\mu_x^2 + \mu_{\hat{x}}^2 + C_1)(\sigma_x^2 + \sigma_{\hat{x}}^2 + C_2)}
\end{equation}
$\mu$, $\sigma^2$, and $\sigma_{x\hat{x}}$ are the mean, variance, and covariance, respectively. SSIM evaluates perceptual similarity considering luminance, contrast, and structure.

\textbf{Bit Retrieval Error (BRE):}
\begin{equation}
\text{BRE} = 1 - \text{acc}
\end{equation}
BRE complements accuracy by indicating the proportion of message bits that were incorrectly retrieved. Lower BRE implies stronger robustness.

\begin{table}[ht]
\centering
\caption{Full names of publication sources used in Table~\ref{table:Comparison of different methods}.}
\label{tab:source_fullnames}
\begin{tabular}{lc}
\toprule
\textbf{Abbreviation} & \textbf{Full Name} \\
\midrule
TNSE   & IEEE Transactions on Network Science and Engineering \\
AAAI   & AAAI Conference on Artificial Intelligence \\
TMM    & IEEE Transactions on Multimedia \\
TDSC   & IEEE Transactions on Dependable and Secure Computing \\
TCSVT  & IEEE Transactions on Circuits and Systems for Video Technology \\
TIFS   & IEEE Transactions on Information Forensics and Security \\
NeurIPS & Conference on Neural Information Processing Systems \\
\bottomrule
\end{tabular}
\end{table}

\section{Motivation and Mathematical Proof}
\label{sec:Mathematical Proof}

\begin{figure}[H]
\centering
\includegraphics[width=3.5in]{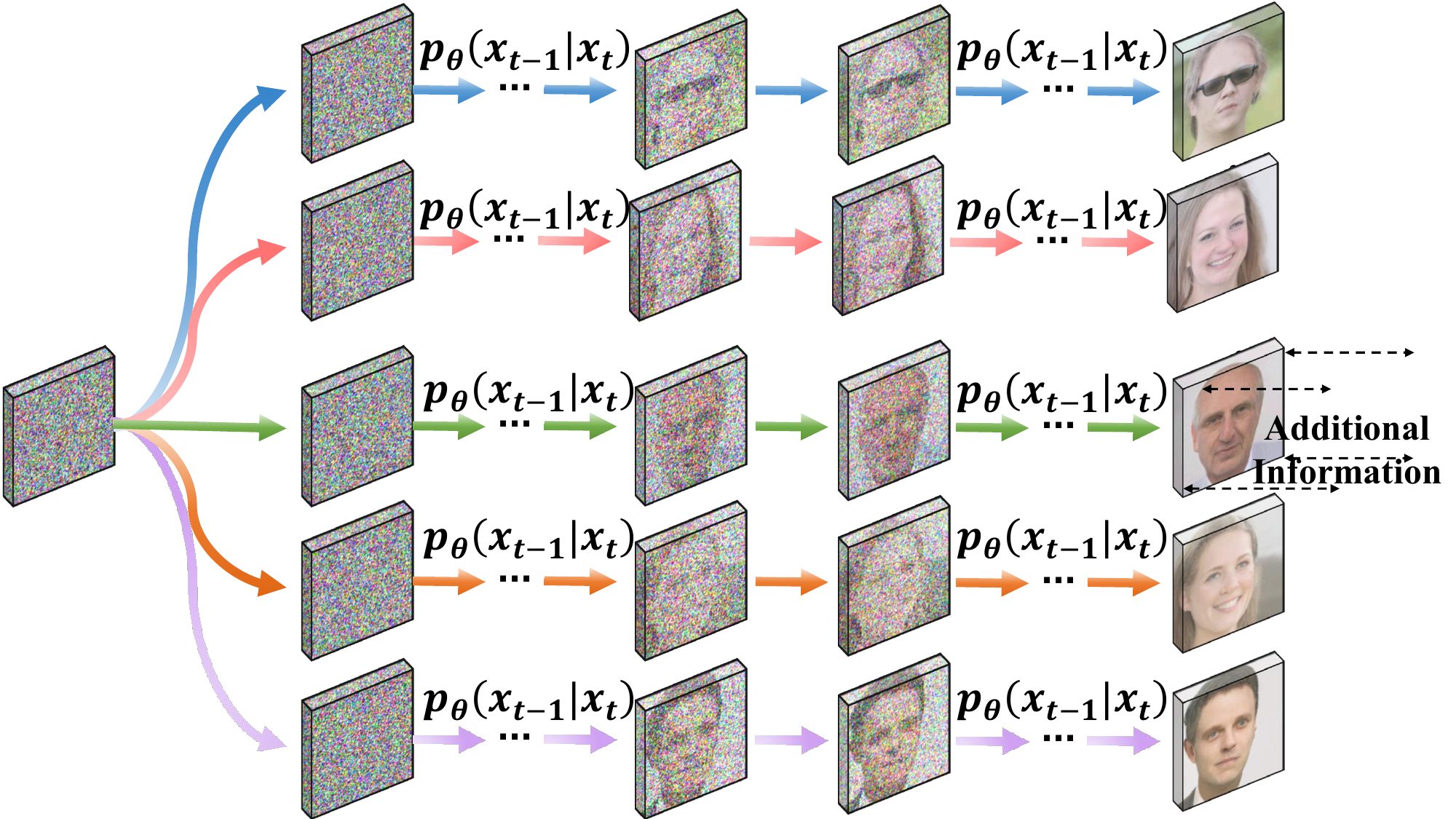}
\caption{Potential Steganographic Targets in Diffusion Model Generation.}
\label{fig:Potential Steganographic Targets}
\end{figure}

It is important to emphasize that the image generation process in diffusion models exhibits a one-to-many mapping, wherein a single Gaussian distribution can give rise to multiple target images (the final meaningful outputs). This characteristic underscores the model’s ability to produce a diverse range of semantically coherent results from a single probabilistic input, highlighting its capacity to capture variability and represent the complexity of image distributions.

The integration of stochastic processes within diffusion models yields a crucial insight: among the vast array of potential outputs that conform to the empirical distribution of complex images, there exists a subset that adheres to specific constraints. These constraints are such that the bit values at designated pixel locations remain fixed. As illustrated in Fig. \ref{fig:Potential Steganographic Targets}, our analysis leads to a compelling conclusion: within the spectrum of meaningful images generated by diffusion models, certain outputs may subtly embed additional information. This hidden data is seamlessly integrated into the image's structure, becoming an intrinsic part of it, such that no overt anomalies are detectable upon superficial inspection of the generated image

We formalize the feasibility of embedding binary information into the lower bits of pixels during selected denoising steps of DDPM sampling, without compromising the quality or semantic consistency of the generated image.

\paragraph{Preliminaries.}
Let $\{x_T, x_{T-1}, \dots, x_0\}$ denote the reverse process of a Denoising Diffusion Probabilistic Model (DDPM), where the initial sample $x_T \sim \mathcal{N}(0, I)$ is gradually denoised to yield $x_0 \sim p_{\theta}(x_0)$. The reverse sampling step is defined as:

\begin{equation}
x_{t-1} = \frac{1}{\sqrt{\alpha_t}} \left( x_t - \frac{1 - \alpha_t}{\sqrt{1 - \bar{\alpha}_t}} \cdot \epsilon_{\theta}(x_t, t) \right) + \sigma_t z,\quad z \sim \mathcal{N}(0, I)
\end{equation}

\paragraph{Bitwise Embedding as Perturbation.}
Let $m \in \{0,1\}^K$ be the secret message to embed, and $\Lambda \subset [1,H] \times [1,W]$ denote the set of spatial locations selected for embedding via a pseudo-random map $\mathcal{I}: [1,K] \to \Lambda$. We assume sparse embedding: $|\Lambda|/(H \cdot W) < \rho$ for a small $\rho$ (e.g., $\rho < 0.05$).

Let $x_k$ be the intermediate sample at timestep $k$, and define the perturbation:

\begin{equation}
x_k' = x_k + \Delta_k,\quad \Delta_k = \mathcal{Q}_m(x_k) - x_k
\end{equation}

where $\mathcal{Q}_m$ is an operator that replaces the least significant 4 bits (LSB-4) of pixel values in $x_k$ at locations $\Lambda$ with message bits. By construction, $\|\Delta_k\|_\infty \leq 15$ and $\|\Delta_k\|_2 \leq 15 \sqrt{K}$.

\paragraph{Stability of the Generative Process.}
Let $\Phi_{\theta}^{(k \to 0)}: \mathbb{R}^{H \times W} \rightarrow \mathbb{R}^{H \times W}$ denote the deterministic sampling trajectory from $x_k$ to $x_0$ under the learned DDPM reverse process. Assume:

\begin{assumption}[Lipschitz Continuity]
There exists a constant $L > 0$ such that for any perturbed $x_k'$:
\begin{equation}
\|\Phi_{\theta}^{(k \to 0)}(x_k') - \Phi_{\theta}^{(k \to 0)}(x_k)\|_2 \leq L \cdot \|\Delta_k\|_2
\end{equation}
\end{assumption}

Hence, for sufficiently small $\Delta_k$ (controlled via sparse and low-amplitude LSB embedding), the final sample $x_0'$ remains within an $\epsilon$-neighborhood of the clean sample $x_0$:

\begin{equation}
\|x_0' - x_0\|_2 \leq L \cdot 15 \sqrt{K} = \epsilon
\end{equation}

where $\epsilon$ is a visually imperceptible distortion bound.

\paragraph{Message Recoverability.}
Let $\mathcal{D}: \mathbb{R}^{H \times W} \rightarrow \{0,1\}^K$ be the LSB-4 decoder that extracts the message from positions $\Lambda$ in $x_0'$. Define the bitwise recovery accuracy as:

\begin{equation}
\text{Acc}(m, \mathcal{D}(x_0')) = \frac{1}{K} \sum_{i=1}^K \mathbf{1}\{m_i = \mathcal{D}(x_0')[i]\}
\end{equation}

We aim for:

\begin{equation}
\mathbb{P}[\text{Acc}(m, \mathcal{D}(x_0')) \geq \alpha] \geq 1 - \delta
\end{equation}

for some high $\alpha \in [0.95, 1]$ and small $\delta \ll 1$, indicating reliable information recovery.

\paragraph{Conclusion.}
Given:
\begin{itemize}
  \item Perturbation $\|\Delta_k\|_\infty \leq 15$ and sparse embedding $\rho \ll 1$,
  \item Lipschitz continuity of $\Phi_{\theta}^{(k \to 0)}$,
  \item Spatial diffusion of embedding positions via a chaotic pseudo-random map,
\end{itemize}
then:

\begin{equation}
\Phi_{\theta}^{(k \to 0)}(x_k') \in \mathcal{M}_{\epsilon}(x_0) \cap \mathcal{S}_m
\end{equation}

where $\mathcal{M}_{\epsilon}(x_0)$ is the set of perceptually similar images, and $\mathcal{S}_m$ denotes the set of images embedding message $m$ at designated LSB positions. Therefore, low-bit embedding during DDPM sampling is theoretically feasible under controlled conditions.

\section{More Experimental Details and Further Analysis}

\begin{figure}[H]
\centering

\begin{subfigure}[b]{0.12\textwidth}
\includegraphics[width=\textwidth]{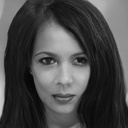}
\caption{}
\end{subfigure}
\hfill
\begin{subfigure}[b]{0.2\textwidth}
\includegraphics[width=\textwidth]{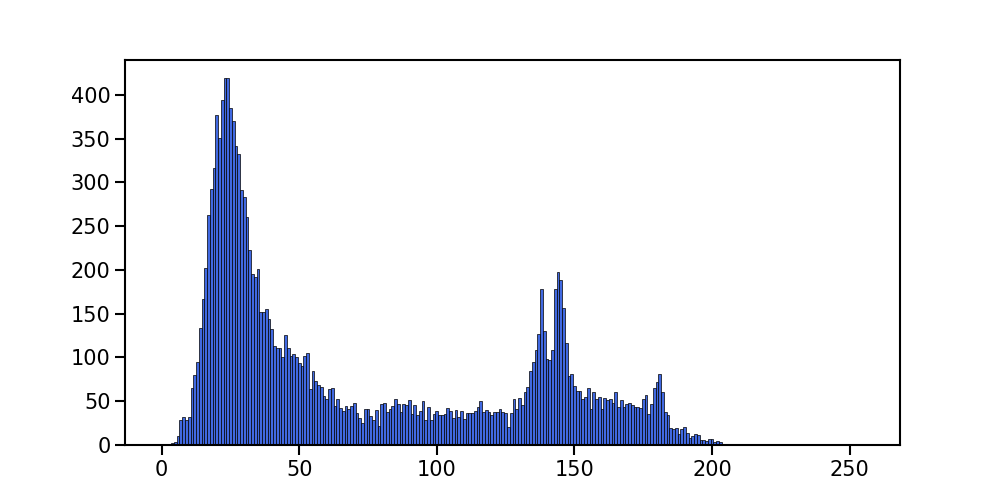}
\caption{}
\end{subfigure}
\hfill
\begin{subfigure}[b]{0.16\textwidth}
\includegraphics[width=\textwidth]{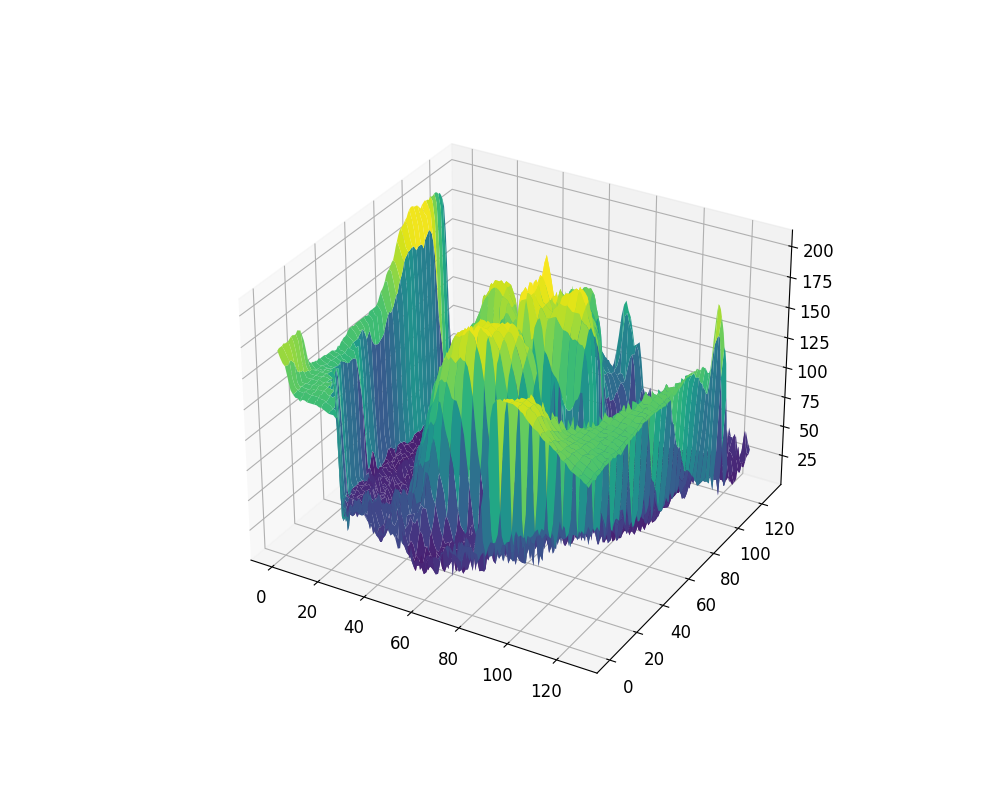}
\caption{}
\end{subfigure}
\hfill
\begin{subfigure}[b]{0.12\textwidth}
\includegraphics[width=\textwidth]{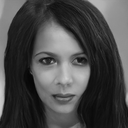}
\caption{}
\end{subfigure}
\hfill
\begin{subfigure}[b]{0.2\textwidth}
\includegraphics[width=\textwidth]{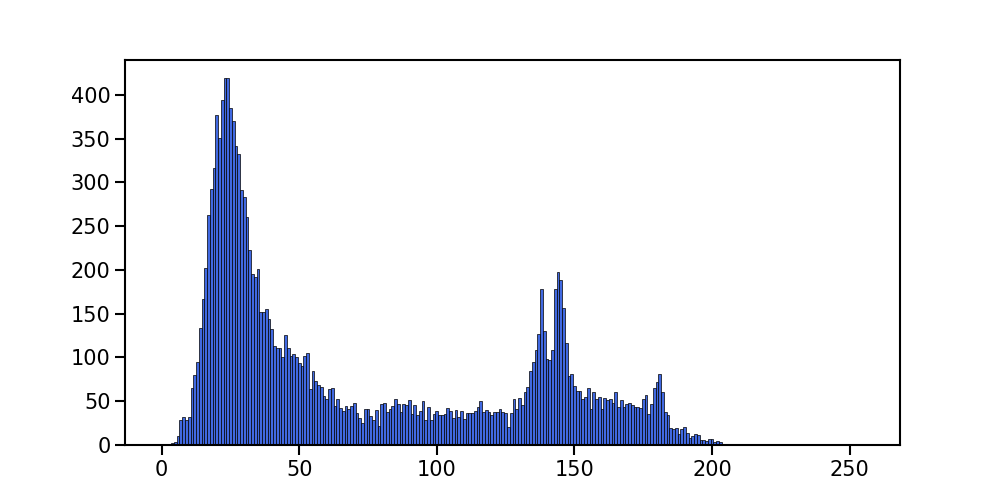}
\caption{}
\end{subfigure}
\hfill
\begin{subfigure}[b]{0.16\textwidth}
\includegraphics[width=\textwidth]{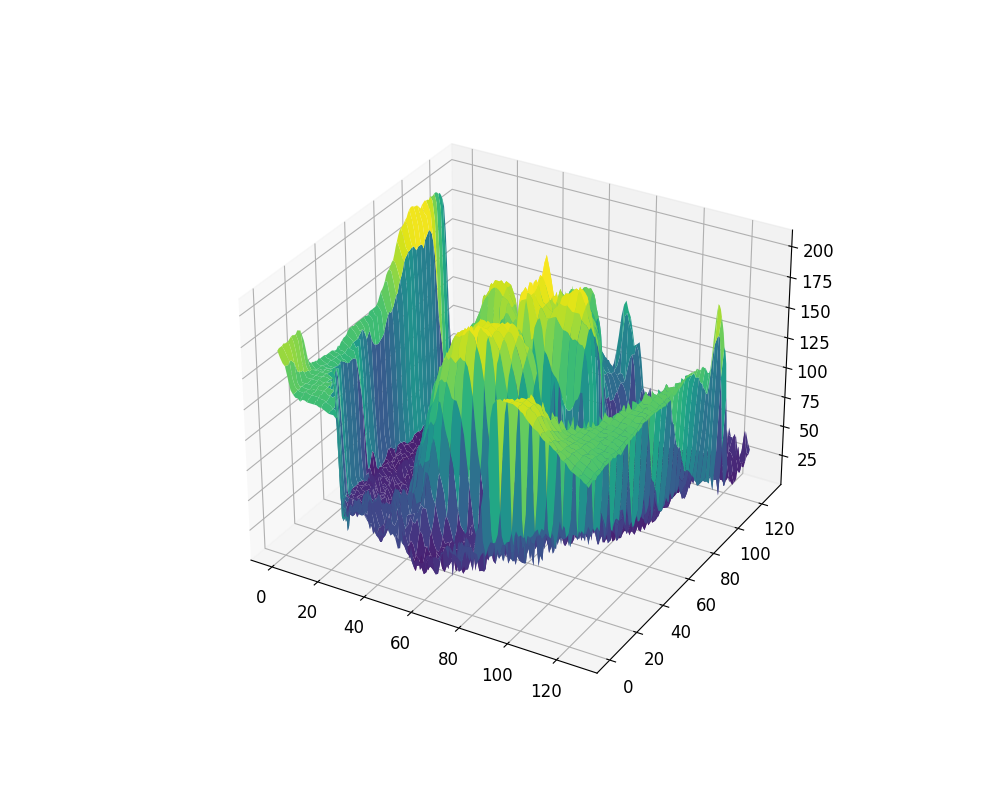}
\caption{}
\end{subfigure}

\caption{Pre- and Post-Steganographic Embedding and Extraction Comparison. (a) shows the image to be steganographed, (b) is the 3D visualization of (a), (c) presents the pixel histogram of (a), (d) represents the extracted image, (e) is the 3D visualization of (d), and (f) shows the pixel histogram of (d).}
\label{fig:Pre and Post}
\end{figure}

As shown in the Figure \ref{fig:Pre and Post}, we present a comparison of the stego image before and after information extraction, including the original image, histograms, and 3D visualization results. From these comparisons, it is clear that the target image extracted from the carrier image remains highly consistent with the original image in terms of visual appearance. Furthermore, histogram and 3D visualization analyses further confirm that both images exhibit identical statistical characteristics. 

This series of evidence strongly demonstrates that the proposed method ensures the integrity and accuracy of the hidden data, achieving the intended goal of complete recoverability. This characteristic is crucial for ensuring the reliability of steganographic applications. 

\begin{figure}[H]
\centering
\begin{subfigure}[b]{0.15\textwidth}
\includegraphics[width=\textwidth]{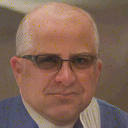}
\caption{}
\end{subfigure}
\hspace{0.05cm} %
\begin{subfigure}[b]{0.15\textwidth}
\includegraphics[width=\textwidth]{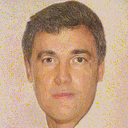}
\caption{}
\end{subfigure}
\hspace{0.05cm} %
\begin{subfigure}[b]{0.15\textwidth}
\includegraphics[width=\textwidth]{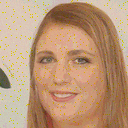}
\caption{}
\end{subfigure}
\hspace{0.05cm} %
\begin{subfigure}[b]{0.15\textwidth}
\includegraphics[width=\textwidth]{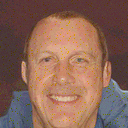}
\caption{}
\end{subfigure}
\hspace{0.05cm} %
\begin{subfigure}[b]{0.15\textwidth}
\includegraphics[width=\textwidth]{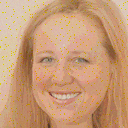}
\caption{}
\end{subfigure}

\vspace{0.05cm} 

\begin{subfigure}[b]{0.15\textwidth}
\includegraphics[width=\textwidth]{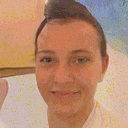}
\caption{}
\end{subfigure}
\hspace{0.05cm} %
\begin{subfigure}[b]{0.15\textwidth}
\includegraphics[width=\textwidth]{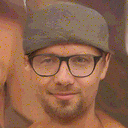}
\caption{}
\end{subfigure}
\hspace{0.05cm} %
\begin{subfigure}[b]{0.15\textwidth}
\includegraphics[width=\textwidth]{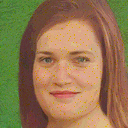}
\caption{}
\end{subfigure}
\hspace{0.05cm} %
\begin{subfigure}[b]{0.15\textwidth}
\includegraphics[width=\textwidth]{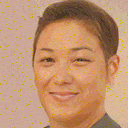}
\caption{}
\end{subfigure}
\hspace{0.05cm} %
\begin{subfigure}[b]{0.15\textwidth}
\includegraphics[width=\textwidth]{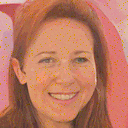}
\caption{}
\end{subfigure}

\caption{The figures showcase our method's effectiveness on multi-modal steganography. (a)-(e) display images generated with embedded text, maintaining visual quality. (f)-(j) show similar results but with embedded audio.}
\label{fig:multi-modal steganography}
\end{figure}

By converting audio information or image information into binary data and then transforming it into a uniform binary distribution as described in Section \ref{sec:Method}, this method enables the embedding of steganographic information in various forms. Figure \ref{fig:multi-modal steganography} illustrates the embedding results of multimodal steganographic information under 4bpp conditions. Taking text information as an example, when English letters or numbers are represented using ASCII encoding, each character occupies 8 bits. Therefore, under the premise of maintaining the naturalness and meaningfulness of the carrier image, a $128 \times 128$ resolution image can carry up to 24,576 English letters or numbers. This finding indicates that the method is not only applicable to image steganography but also capable of effectively embedding textual and audio information, showcasing its broad application potential and adaptability across different scenarios. Thus, this method demonstrates significant advantages in multimodal data steganography, achieving efficient information hiding while maintaining high visual quality, suitable for a variety of application contexts.

\section{Social Impact}
\label{app:social_impact}

\section*{Social Impact}

\textbf{Positive Impacts.} SD$^2$ introduces a novel, carrier-free approach to steganography that significantly enhances secure communication. By embedding information directly into the generative process of diffusion models, it enables high-capacity and lossless message encoding without relying on natural cover media. This feature is particularly beneficial in contexts requiring privacy and anti-censorship measures, such as for journalists, human rights defenders, and at-risk communities. Furthermore, SD$^2$ represents a substantial advancement in data hiding technology. Its use of the denoising and redundancy-exploiting properties of diffusion models leads to improved flexibility and reliability over traditional spatial- and frequency-domain methods. Notably, the algorithm exhibits strong robustness, allowing perfect message extraction even under severe perturbations such as up to 50\% image cropping. Its ability to support multimodal data—text, audio, and more—by mapping them into a unified binary representation also broadens its applicability across education, cultural preservation, and creative industries, where secure, high-fidelity content generation is valuable.

\textbf{Negative Impacts.} Despite its technical merits, SD$^2$ raises significant concerns about misuse. Its capability to generate visually realistic images containing covert information from scratch could be exploited for unlawful communication, presenting challenges for national security and public safety. Because it avoids using natural carriers, the approach evades detection by conventional steganalysis tools, thereby complicating digital forensic investigations and making illicit use harder to trace. Additionally, the complexity of bit-position selection and timestep scheduling currently required for optimal performance may hinder scalability and operational robustness in uncontrolled environments. This complexity, if mishandled, can lead to decoding failures or system instability. Moreover, the lack of fine-grained semantic control limits interpretability and oversight, increasing the risk of unregulated deployment or unintended outcomes. These factors highlight the need for the development of ethical guidelines, detection mechanisms, and regulatory frameworks to ensure the responsible use of generative steganographic methods like SD$^2$.

\end{document}